\documentclass[runningheads]{llncs}

\usepackage[T1]{fontenc}
\def\doi#1{\href{https://doi.org/\detokenize{#1}}{\url{https://doi.org/\detokenize{#1}}}}
\usepackage{graphicx}
\usepackage{amsmath}
\usepackage{amssymb}
\usepackage{booktabs}
\usepackage{caption}
\usepackage{subcaption}  
\usepackage[title]{appendix}
\graphicspath{ {./images/} }
\usepackage{listings}
\usepackage[pagebackref,breaklinks,colorlinks]{hyperref}

 \pdfoutput=1

\begin{document}

\pdfoutput=1

\title{TTS-GAN: A Transformer-based Time-Series Generative Adversarial Network}
\titlerunning{TTS-GAN}
%
\author{Xiaomin Li, Vangelis Metsis, Huangyingrui Wang, Anne Hee Hiong Ngu \\
{\tt\small x\_l30, vmetsis, h\_w91, angu @txstate.edu}}
\authorrunning{X. L. et al.}
%
\institute{Texas State University, San Marcos TX 78666, USA}
\maketitle              
\begin{abstract}
Signal measurements appearing in the form of time series are one of the most common types of data used in medical machine learning applications. However, such datasets are often small, making the training of deep neural network architectures ineffective. For time-series, the suite of data augmentation tricks we can use to expand the size of the dataset is limited by the need to maintain the basic properties of the signal. Data generated by a Generative Adversarial Network (GAN) can be utilized as another data augmentation tool. RNN-based GANs suffer from the fact that they cannot effectively model long sequences of data points with irregular temporal relations. To tackle these problems, we introduce TTS-GAN, a transformer-based GAN which can successfully generate realistic synthetic time-series data sequences of arbitrary length, similar to the real ones. Both the generator and discriminator networks of the GAN model are built using a pure transformer encoder architecture. We use visualizations and dimensionality reduction techniques to demonstrate the similarity of real and generated time-series data. We also compare the quality of our generated data with the best existing alternative, which is an RNN-based time-series GAN.

TTS-GAN source code: \href{https://github.com/imics-lab/tts-gan}{github.com/imics-lab/tts-gan}

\keywords{Generative Adversarial Network \and Transformer \and Time-Series Analysis \and Medical Signal }
\end{abstract}
\section{Introduction}
\label{sec:intro}
Data shortage is often an issue when analyzing physiology based time-series signals with deep learning models. Unlike images and text data used in computer vision (CV) and natural language processing (NLP) tasks, which are abundant on the web, such signals are collected as sensor measurements resulting from physical or biological process. Especially when such processes involve human subjects, data collection, annotation, and interpretation is a costly endeavour. Furthermore, differences in the various collection configurations make it harder for data collected in different settings to be merged together to form larger datasets. Deep learning models require large amounts of data to train successfully. Training deep learning models with a high number of trainable parameters on small datasets results in over-fitting and low generalization capabilities. As a compromise researchers are forced to train shallower deep learning models that are not capable of capturing the full complexity of the problem at hand. This is a common situation encountered in medical and health-related machine learning research.

Generative Adversarial Networks (GANs), first introduced in 2014~\cite{goodfellow2014generative}, have been gaining traction in the deep learning research field. They have successfully generated and manipulated data in CV and NLP domains, such as high-quality image generation~\cite{ledig2017photo}, style transfer~\cite{bousmalis2017unsupervised}, text-to-image synthesis~\cite{zhang2017stackgan}, etc. There has also been a movement towards using GANs for time series and sequential data generation, and forecasting. The review paper~\cite{brophy2021generative} gives a thorough summary of GAN implementations on time series data. 

A GAN is a generative model consisting of a generator and discriminator, typically two neural network (NN) models. The generator takes as input random vectors of specified dimensions and generates output vectors of the same dimension that are similar to the real training data. The discriminator is a binary classifier used to distinguish the real data and generated data. The generator and discriminator are updated by back-propagation alternately, playing a zero-sum game against each other and until they reach an equilibrium.  

The transformer architecture, which relies on multiple self-attention layers~\cite{vaswani2017attention}, has recently become a prevalent deep learning model architecture. It has been shown to surpass many other popular neural network architectures, such as CNN over images and RNN over sequential data~\cite{dosovitskiy2020image,devlin2018bert}, and it has even displayed properties of a universal computation engine~\cite{lu2021pretrained}. Some works have already tried to utilize the transformer model in GAN model architecture design with the goal to either improve the quality of synthetic data or to create a more efficient training process~\cite{jiang2021transgan,diao2021tilgan} for image and text generation tasks. In work~\cite{jiang2021transgan}, the author, for the first time, built a pure transformer-based GAN model and verified its performance on multiple image synthesis tasks. 

Previous efforts for creating a time-series GAN have mainly relied on Recurrent Neural Network (RNN)-based architectures~\cite{esteban2017real,yoon2019time,ni2020conditional}. Since the transformer was first invented to handle very long sequential data and does not suffer from a vanishing gradient problem, theoretically, a transformer GAN model should perform better than other RNN-based models on time-series data. In this work, we follow a process similar to the one Jiang et.al.~\cite{jiang2021transgan} followed for image generation, adapted for time-series data. 

Since time-series data are not easily interpretable by humans, we use PCA~\cite{wold1987principal} and t-SNE~\cite{van2008visualizing} to map the multi-dimensional output sequence vectors into two dimensions to visually observe the similarity in the distribution of the synthetic data and real data instances. For a more quantitative comparison, we also measure several well-known signal properties and compare the similarity of the transformer-generated as well as RNN-generated sequences with real sequences of the same class.



Our contributions can be summarized as follows:
\begin{itemize}
    \item We create a pure transformer-based GAN model to generate synthetic time-series data.
    \item We propose several heuristics to more effectively train a transformer-based GAN model on time-series data. 
    \item We qualitatively and quantitatively compare the quality of the generated sequences against real ones and against sequences generated by other state-of-the-art time-series GAN algorithms. 
\end{itemize}

The rest of the paper is organized as follows. Section~\ref{sec:background} discusses the background and most popular applications of GANs and transformer models. In section~\ref{sec:methodology}, we provide the details of our TTS-GAN model architecture and how we process time-series data to feed this model. In section~\ref{sec:experiments}, we visually and quantitatively verify the fidelity of the synthetic data. 
Section~\ref{sec:conclude} summarizes our work and concludes this paper.

\section{Background}
\label{sec:background}

\subsection{Generative Adversarial Networks (GANs)}
GANs consist of two models, a generator and a discriminator. These two models are typically implemented by neural networks, but they can be implemented with any form of differentiable system that maps data from one space to the other. The generator tries to capture the distribution of true examples for new data example generation. The discriminator is usually a binary classifier, discriminating generated examples from the true examples as accurately as possible. The optimization of GANs is a minimax optimization problem, in which the goal is to reach Nash equilibrium~\cite{ratliff2013characterization} of the generator and discriminator. Then, the generator can be thought to have captured the real distribution of true examples. 

GANs have had many applications in different areas, but mostly in CV and NLP. For example, it can generate examples for image datasets~\cite{goodfellow2020generative},  front view faces~\cite{huang2017beyond}, text-to-image translation~\cite{zhang2017stackgan}, etc. While these successes have drawn much attention, GAN applications have diversified across disciplines such as time-series data generation. The work~\cite{brophy2021generative} gives a thorough summary of the GAN implementations in this field. The applicability of GANs to this type of data can solve many issues that current dataset holders face. For example, GANs can augment smaller datasets by generating new, previously unseen data. GANs can replace the artifacts with information representative of clean data. And it can also be used to denoise signals. GANs can also ensure an extra layer of data protection by generating deferentially private datasets containing no risk of linkage from source to generated datasets.

\subsection{Transformer}
The transformer is the state-of-the-art neural network architecture. Unlike recurrent neural networks, which consume a sequence token by token, in a transformer network, the entire sequence is fed into layers of transformer modules. The representation of a token at a layer is then computed by attending to the latent representations of all the other tokens in the preceding layer. Many works in the NLP field have proved its performance~\cite{vaswani2017attention,devlin2018bert}. 

Given its strong representation capabilities, researchers have also applied transformers to computer vision tasks. In a variety of visual benchmarks, transformer models perform similar to or better than other types of networks, such as convolutional and recurrent networks. The work in~\cite{dosovitskiy2020image} builds a model named ViT, which applies a pure transformer directly to sequences of image patches. The work in~\cite{jiang2021transgan} builds a pure transformer GAN model to generate synthetic images, where the discriminator designing idea is from the ViT model. The multi-dimension time-series data we are dealing with has similarities from both texts and images, meaning a sequence contains both temporal and spatial information. Each timestep in a sequence is like a pixel on one image. The whole sequence contains an event or multiple events happening, which is similar to a sentence in NLP tasks. 

In this work, we adapt the ideas used in~\cite{dosovitskiy2020image} and~\cite{jiang2021transgan} for images, and view a time-series sequence as a $C \times H \times W $ tuple, where \emph{C} is the number of channels of the time-series data, \emph{H} corresponds to the height of the image, but for time-series that value is set to 1, and\emph{W} corresponds to the width of the image, which for times-series is the number of timesteps in the sequence. 
We divide the tuple into multiple patches on the \emph{W} axis and provide positional encoding to each patch. To our best knowledge, it is the first work to implement such an idea to process time-series data and apply it to a transformer GAN model.

\section{Methodology}
\label{sec:methodology}

\subsection{Transformer Time-Series GAN Model Architecture}

\begin{figure}[t]
\centering
\includegraphics[width=9cm, height=9.5cm]{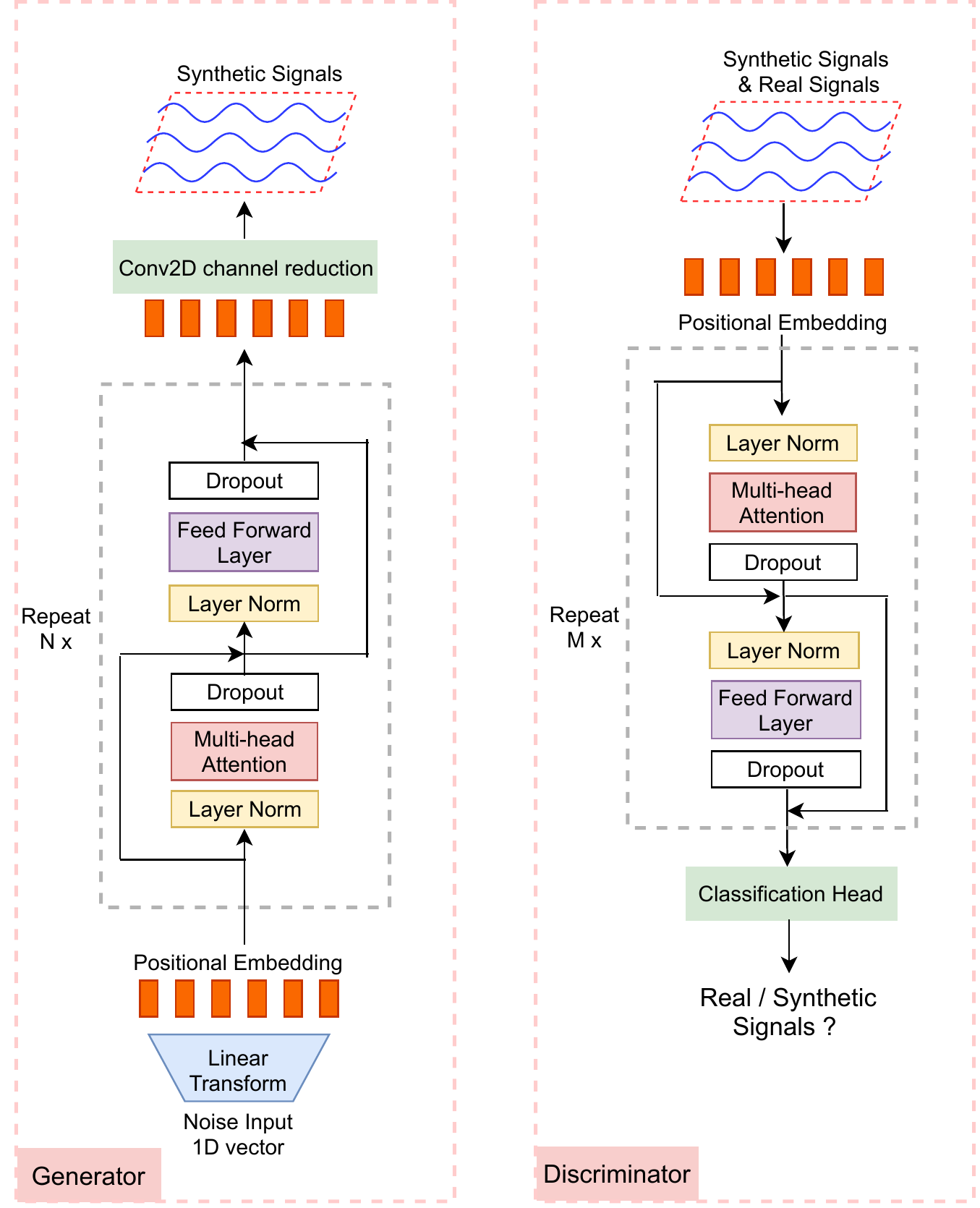}
\caption{TTS-GAN model architecture}
\label{fig:GAN}
\end{figure}

The TTS-GAN model architecture is shown in Fig.~\ref{fig:GAN}. It contains two main components, a generator, and a discriminator. Both of them are built based on the transformer encoder architecture~\cite{vaswani2017attention}. An encoder is a composition of two compound blocks. A multi-head self-attention module constructs the first block and the second block is a feed-forward MLP with GELU activation function. The normalization layer is applied before both of the two blocks and the dropout layer is added after each block. Both blocks employ residual connections. 

The generator first takes in a 1D vector with N uniformly distributed random numbers values within the range (0,1), i.e. $N_i \sim U(0,1)$ . N represents the latent dimension of the synthetic signals, which is a hyperparameter that can be tuned. The vector is then mapped to a sequence with the same length of the real signals and M embedding dimensions. M is also a hyperparameter that can be changed and not necessarily equal to real signal dimensions. Next, the sequence is divided into multiple patches, and a positional encoding value is added to each patch. Those patches are then input to the transformer encoder blocks. Then the encoder blocks outputs are passed through a Conv2D layer to reduce the synthetic data dimensions. The Conv2D layer is set to have a kernel size $(1, 1)$, which won't change the width and height of the synthetic data. The filter size is set to the same dimension size as the real data sequences. Therefore, a synthetic data sequence after the generator transformer encoder layers with a data shape $(hidden dimensions, 1, timesteps)$ will be mapped to $(real data dimensions, 1, timesteps)$. In this way, a random noise vector is transformed into a sequence with the same shape as the real signals. 

The discriminator architecture is similar to the ViT model~\cite{dosovitskiy2020image}, which is a binary classifier to distinguish whether the input sequence is a real signal or synthetic one. In the ViT model, an image is divided evenly into multiple patches with the same width and height. However, in TTS-GAN, we view any input sequences like an image with a height of 1. The timesteps of the inputs are image widths. Therefore, to add positional encoding on time series inputs, we only need to divide the width evenly into multiple pieces and keep the height of each piece unchanged. This process is explained in detail in section~\ref{sec:processingdata}. 

\subsection{Processing Time-Series Data like an image}
\label{sec:processingdata}
\begin{figure}[t]
\centering
\includegraphics[scale=0.7]{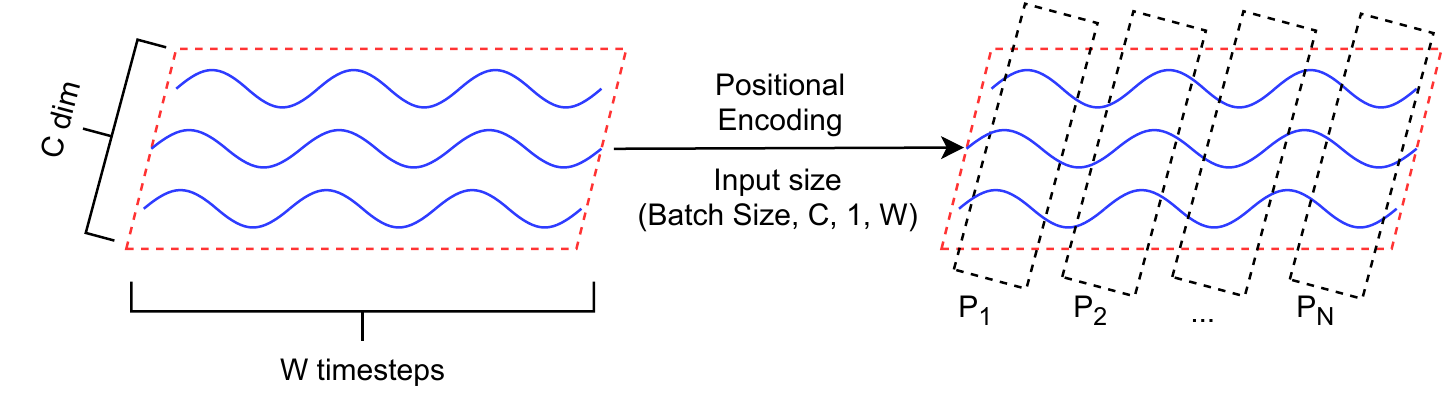}
\caption{Processing time-series data}
\label{fig:PositionalEncoding}
\end{figure}

We view a time-series data sequence like an image with a height equal to 1. The number of timesteps is the width of an image, $W$. A time-series sequence can have a single channel or multiple channels, and those can be viewed as the number of channels (RGB) of an image, $C$. So the input sequences can be represented with the matrix of size $(Batch Size, C, 1, W)$. Then we choose a patch size $N$ to divide a sequence into $W / N$ patches. We then add a soft positional encoding value by the end of each patch, the positional value is learned during model training. Therefore the inputs to the discriminator encoder blocks will have the data shape $(Batch Size, C, 1, (W/N) + 1)$. This process is shown in Fig.~\ref{fig:PositionalEncoding}.

\subsection{Updating Generator and Discriminator Parameters}
\label{sec:loss}
The transformer blocks in the generator and discriminator both use the Mean Squared Error loss to update the parameters. We can use $z$ to denote input vectors to the generator. Use $G(z)$ to represent the synthetic data generated by the generator. We use the prefix $real$ to represent the real input signals. $D(x)$ is the classification output of the discriminator. $x$ can be the real signals or synthetic signals. $real\_label$ is set to 1 and $fake\_label$ is set to 0. To stabilize the GAN model training, some heuristics can be used when setting label values. For example, we can use soft labels that $real\_label$ is a float number close to 1 and $fake\_label$ is a float number close to 0. Sometimes, we can also flip the values of the $real\_label$ and the $fake\_label$. The usefulness of these strategies has been so fat been tested only on a case-by-case basics.
The discriminator loss can be represented as:
\begin{gather*} 
d\_real\_loss = MSELoss(D(real), real\_label) \\
d\_fake\_loss = MSELoss(D(G(z)), fake\_label) \\
d\_loss = d\_real\_loss + d\_fake\_loss 
\end{gather*}
The discriminator loss is the sum of real data loss and fake data (synthetic data) losses. 
The generator loss can be represented as: 
\begin{equation*}
g\_loss = MSELoss(D(G(z)), real\_label)
\end{equation*}

\section{Experiments}
\label{sec:experiments}

\subsection{Datasets}
\begin{figure*}[t]
\centering
\begin{subfigure}{0.495\textwidth}
\centering
\includegraphics[width=\columnwidth]{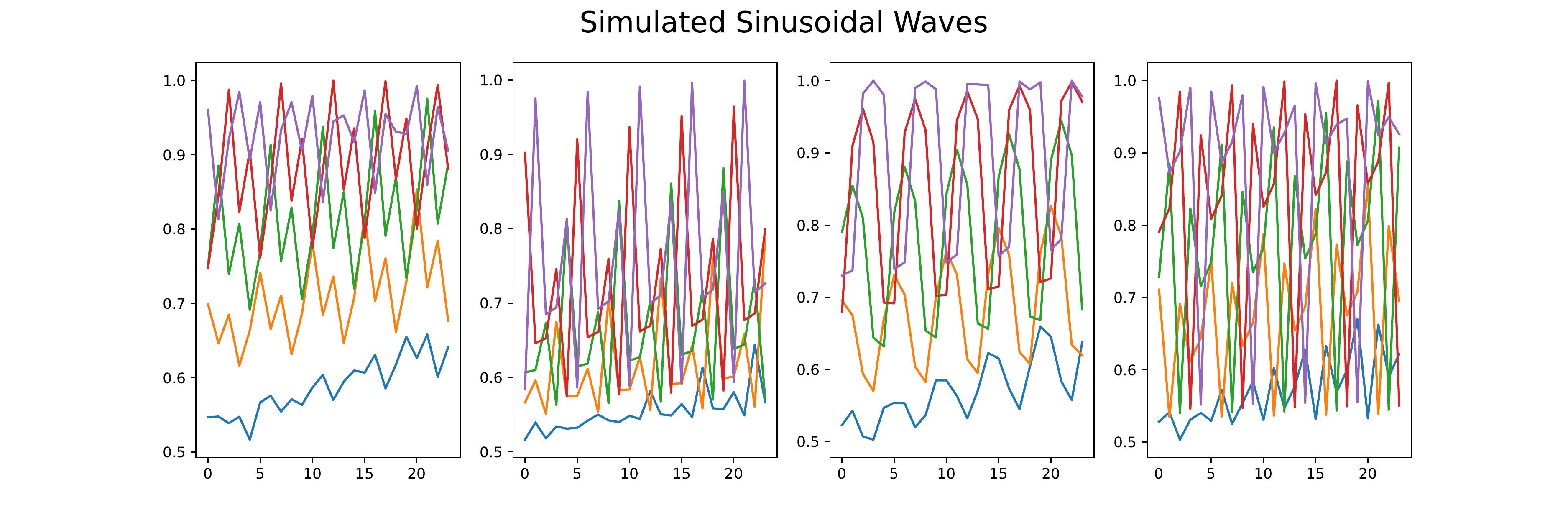} \\
\includegraphics[width=\columnwidth]{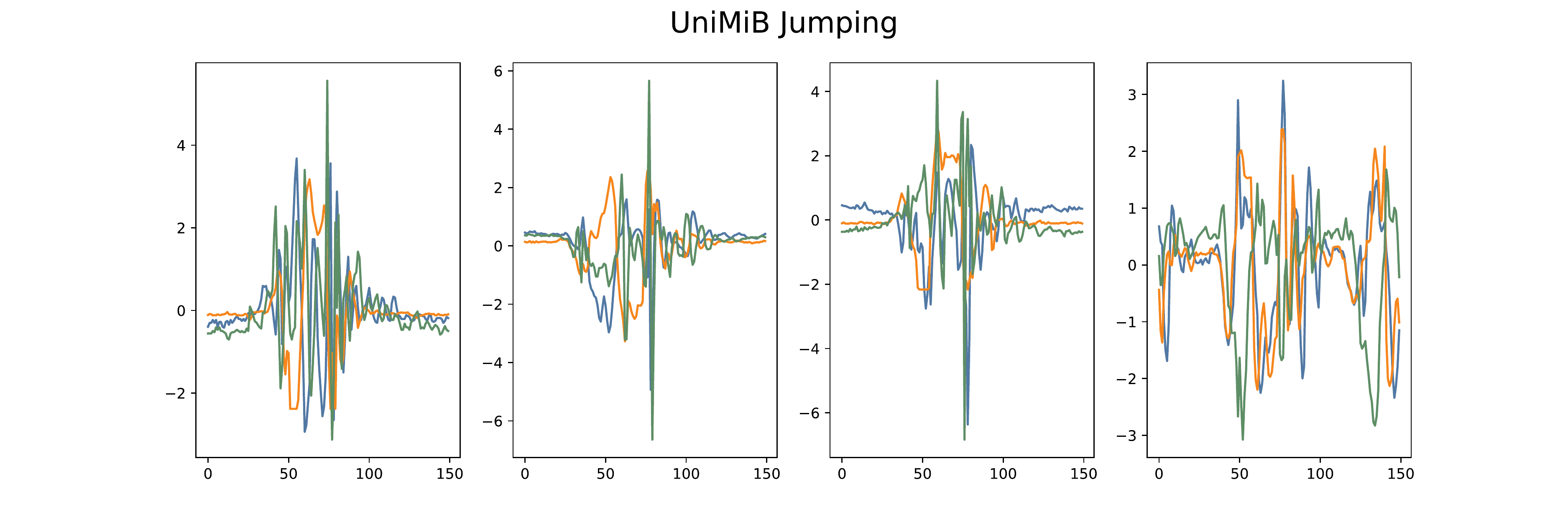} \\
\includegraphics[width=\columnwidth]{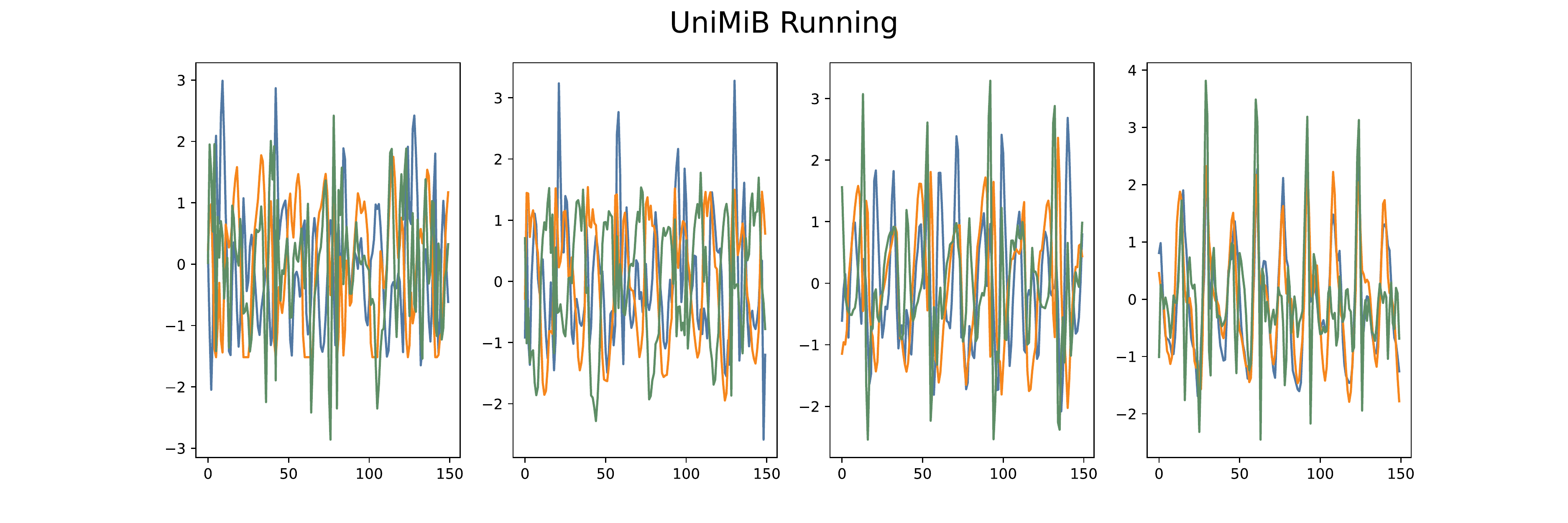} \\
\includegraphics[width=\columnwidth]{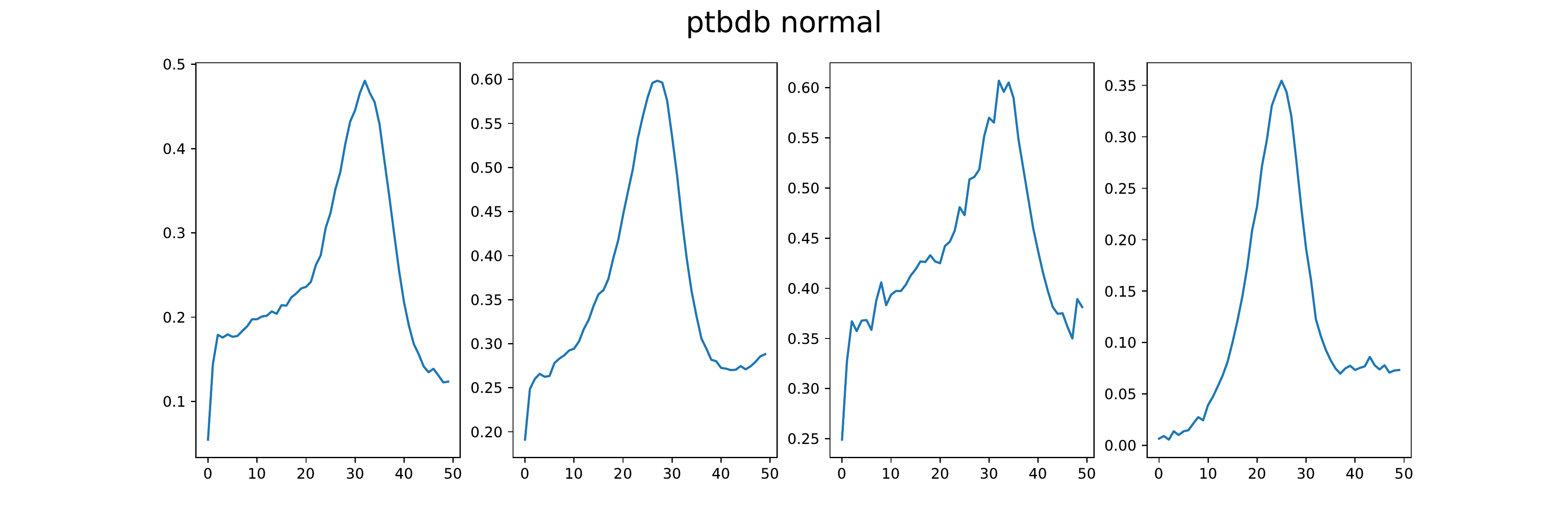} \\
\includegraphics[width=\columnwidth]{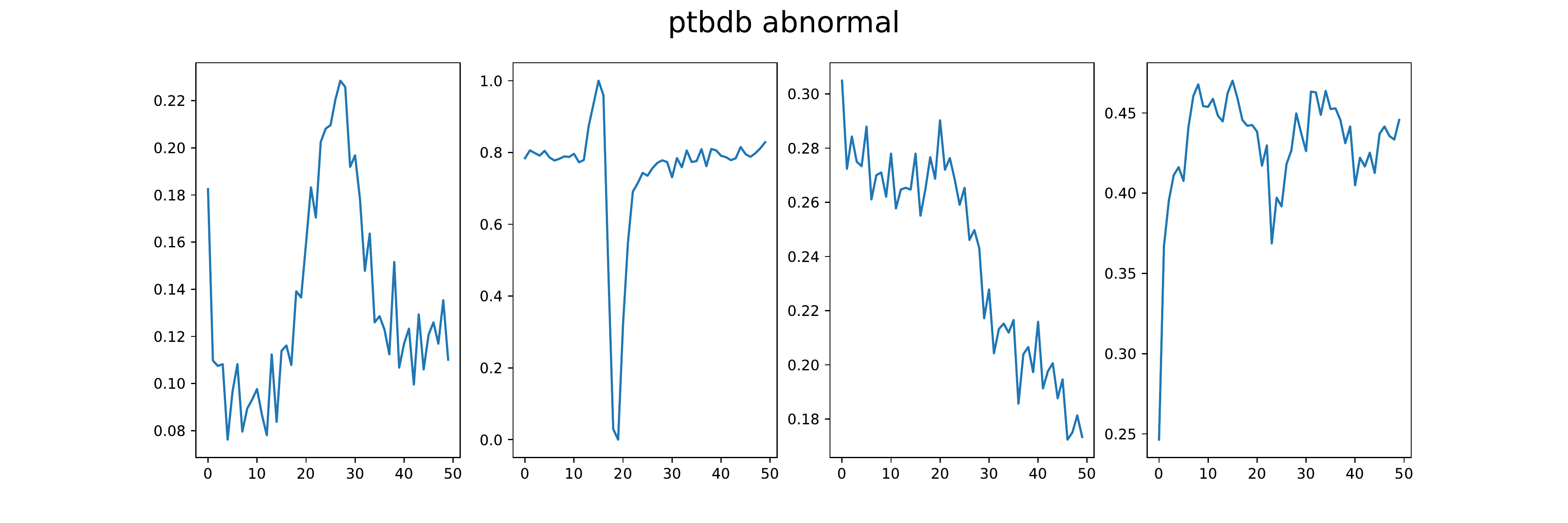} 

\caption{Real signals}
\label{fig:realSignals}
\end{subfigure}
\hfill
\begin{subfigure}{0.495\textwidth}
\centering
\includegraphics[width=\columnwidth]{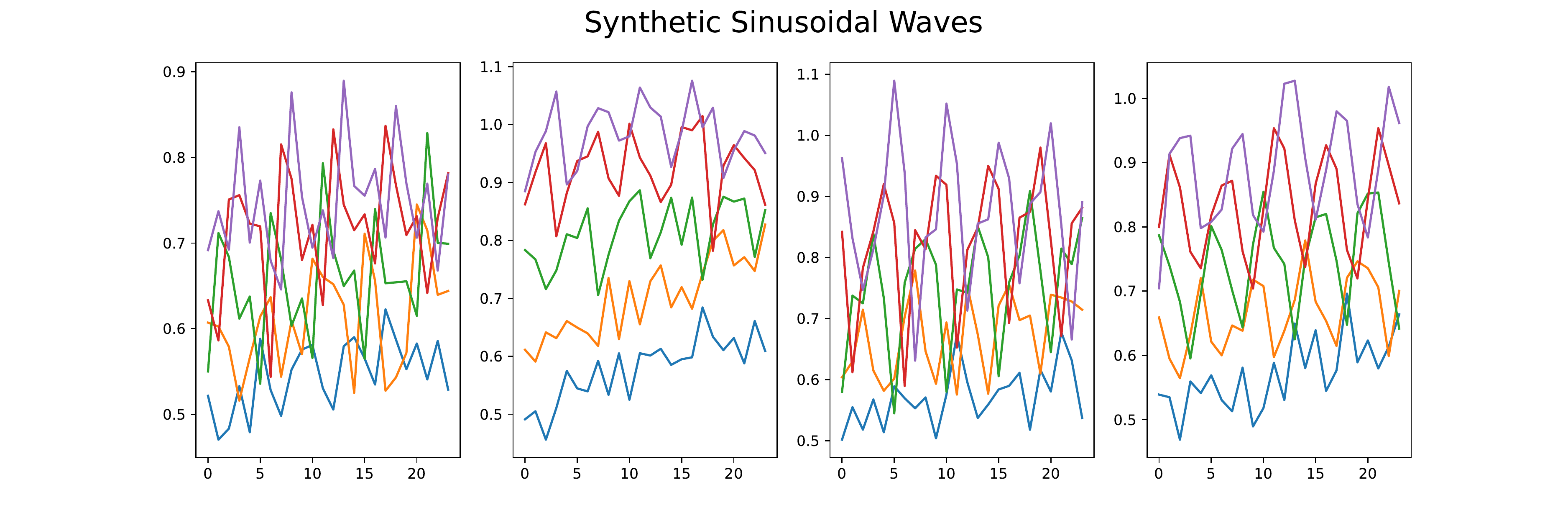} \\
\includegraphics[width=\columnwidth]{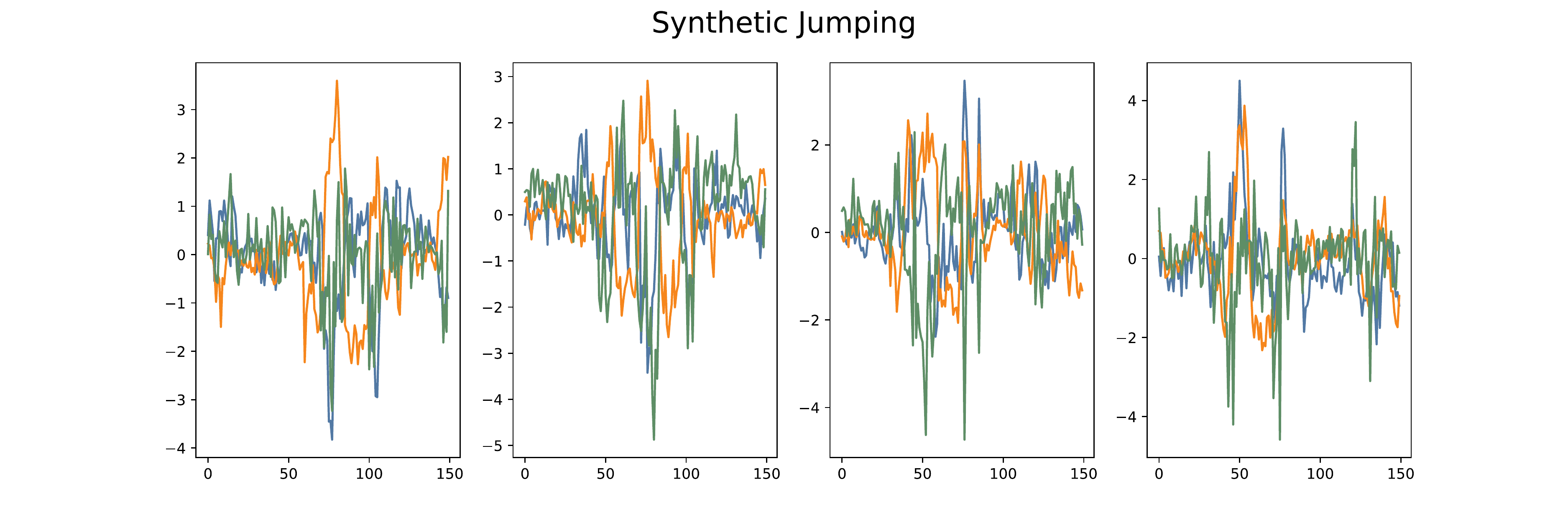} \\
\includegraphics[width=\columnwidth]{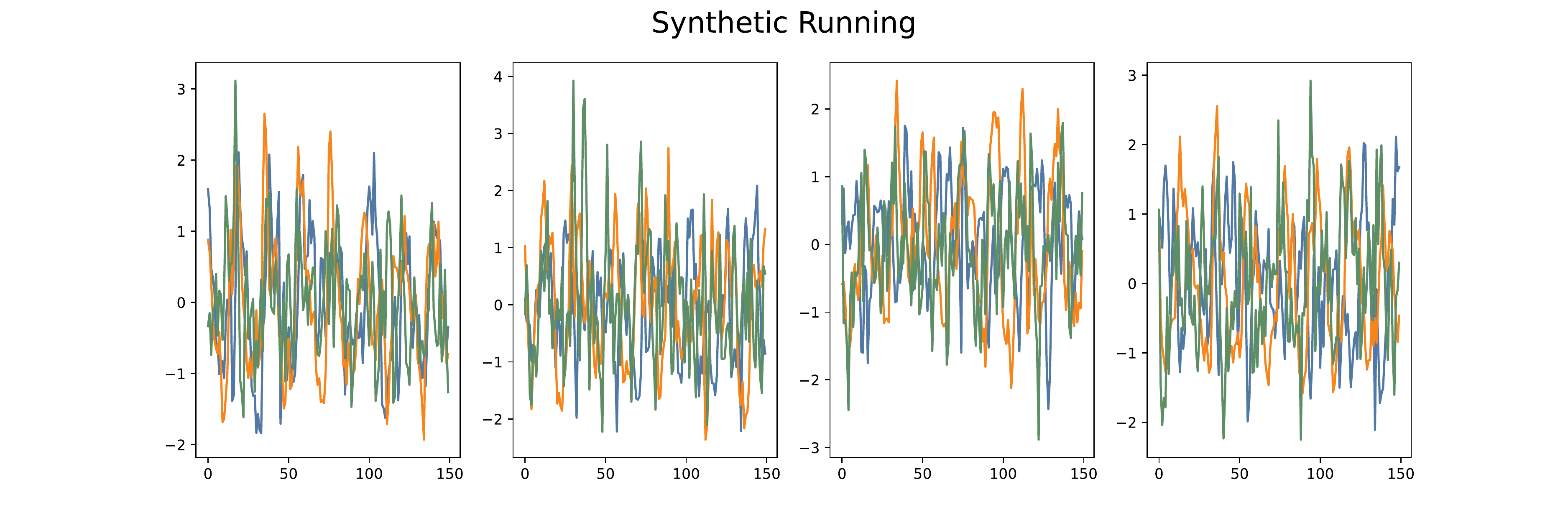} \\
\includegraphics[width=\columnwidth]{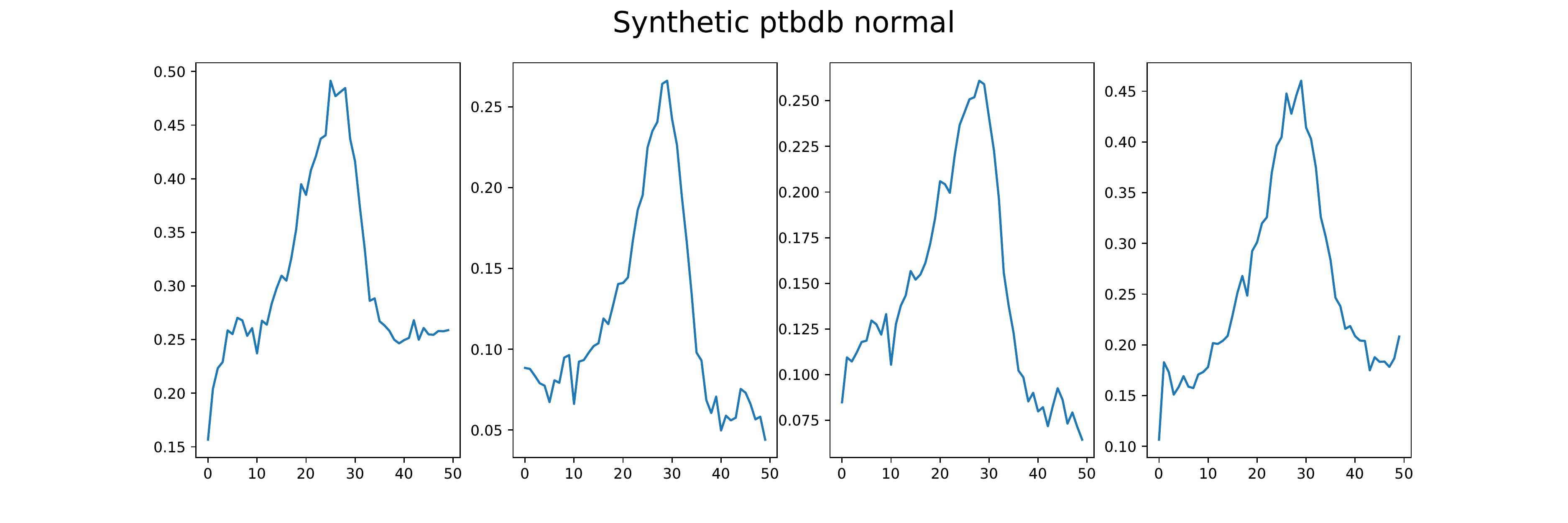} \\
\includegraphics[width=\columnwidth]{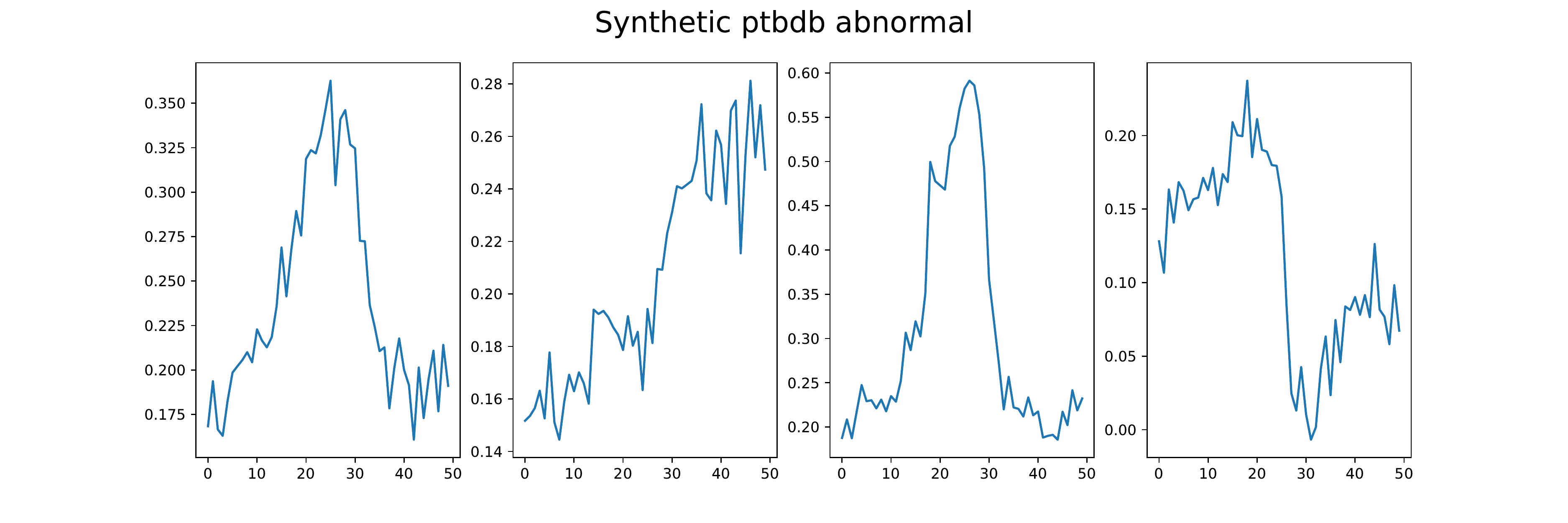} 

\caption{TTS-GAN Synthetic signals}
\label{fig:syntheticSignals}
\end{subfigure}

\caption{A visual comparison of real data and their corresponding synthetic data generated by TTS-GAN.}
\vspace{-5mm}
\end{figure*}

We evaluate the TTS-GAN model on three datasets. Simulated sinusoidal waves,  UniMiB human activity recognition (HAR) dataset~\cite{app7101101} and the PTB Diagnostic ECG Database~\cite{bousseljot1995nutzung,goldberger2000physiobank}. A few raw data samples for each dataset are shown in Fig.~\ref{fig:realSignals}.

The \textbf{sinusoidal waves} are simulated with random frequencies $A$ and phases $B$ values between [0, 0.1]. The sequence length is 24 and the number of dimensions is 5. For each dimension $i \in \{1, ..., 5\}$, the sequence can be represented with the formula $x_i(t) = sin(At + B)$, where $A \in (0, 0.1)$ and $B \in (0, 0.1)$. A total number of 10000 simulated sinusoidal waves are used to train the GAN model.

For the \textbf{UniMiB datase}~\cite{app7101101}, we select 2 categories (Jumping and Running) samples from 24 subjects' recordings to train GAN models. The two classes have 600 and 1572 samples respectively. Every sample has 150 timesteps and three accelerator values at each timestep. All of the recordings are channel-wisely normalized to a mean of 0 and a variance of 1.

The \textbf{PTB Diagnostic ECG dataset}~\cite{bousseljot1995nutzung,goldberger2000physiobank} contains human heartbeat signals in two categories, normal and abnormal with 4046 and 10506 samples respectively. Each sequence represents a heart beat sampled at 125Hz. The original length of each sequence is 188, padded with zeros at the end to create fixed-length sequences. We only use the timesteps 5 to 55 of each sample, which is the part of the sequence containing the most useful information of the heatbeat.

\subsection{Evaluation}

We evaluate TTS-GAN using qualitative visualizations and quantitative metrics, and compare it with Time-GAN~\cite{yoon2019time}, which is the best current alternative.

\noindent \textbf{Raw data visualization:}
Fig.~\ref{fig:syntheticSignals} shows samples of synthetic data generated by TTS-GAN. Comparing them to the real data in Fig.~\ref{fig:realSignals}, we can observe that the synthetic data present visually similar signal patterns to the real data. 

\begin{figure*}
    \begin{subfigure}{0.245\textwidth}
        \centering
        \includegraphics[width=\columnwidth]{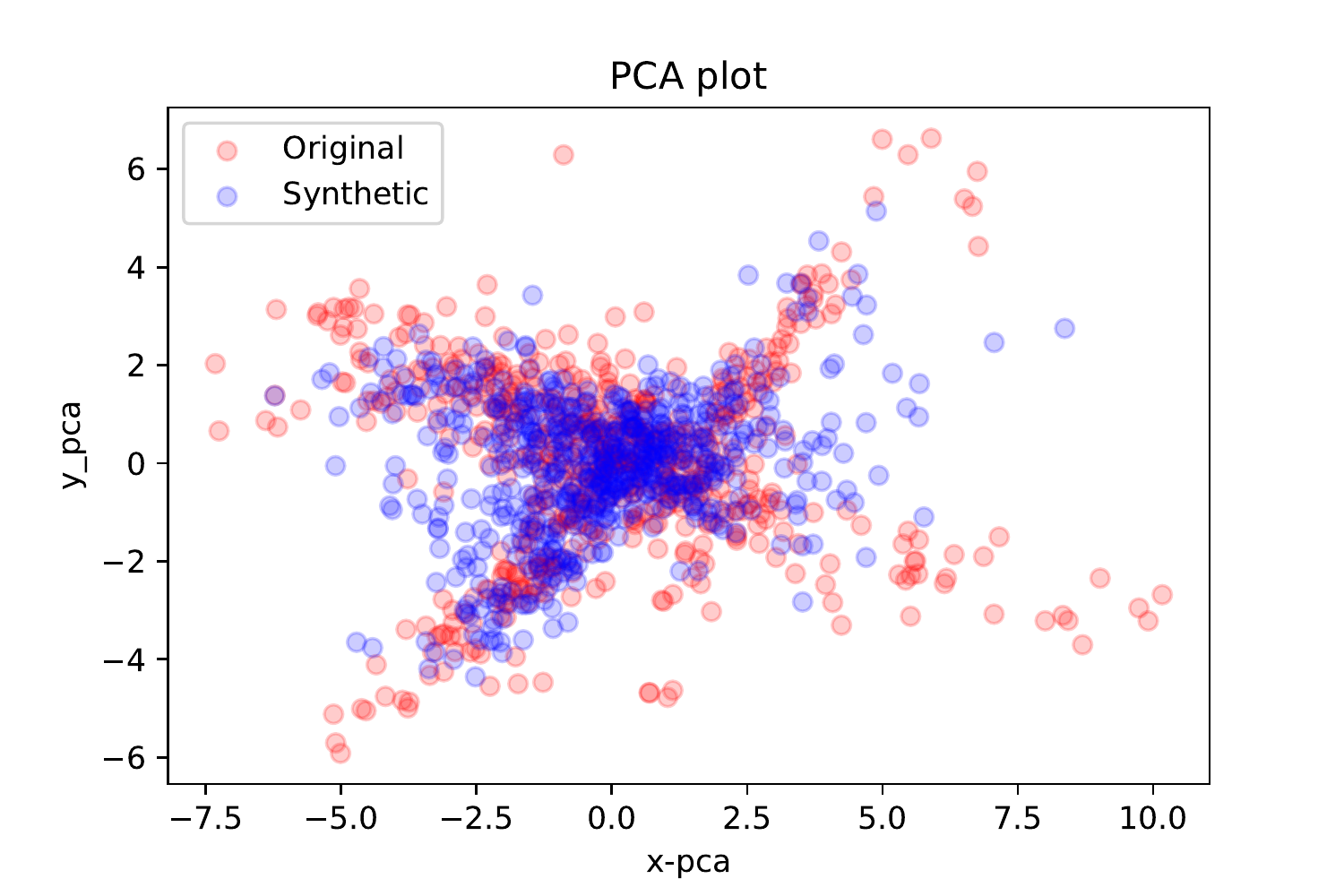}
    \end{subfigure}
    \begin{subfigure}{0.245\textwidth}  
        \centering 
        \includegraphics[width=\columnwidth]{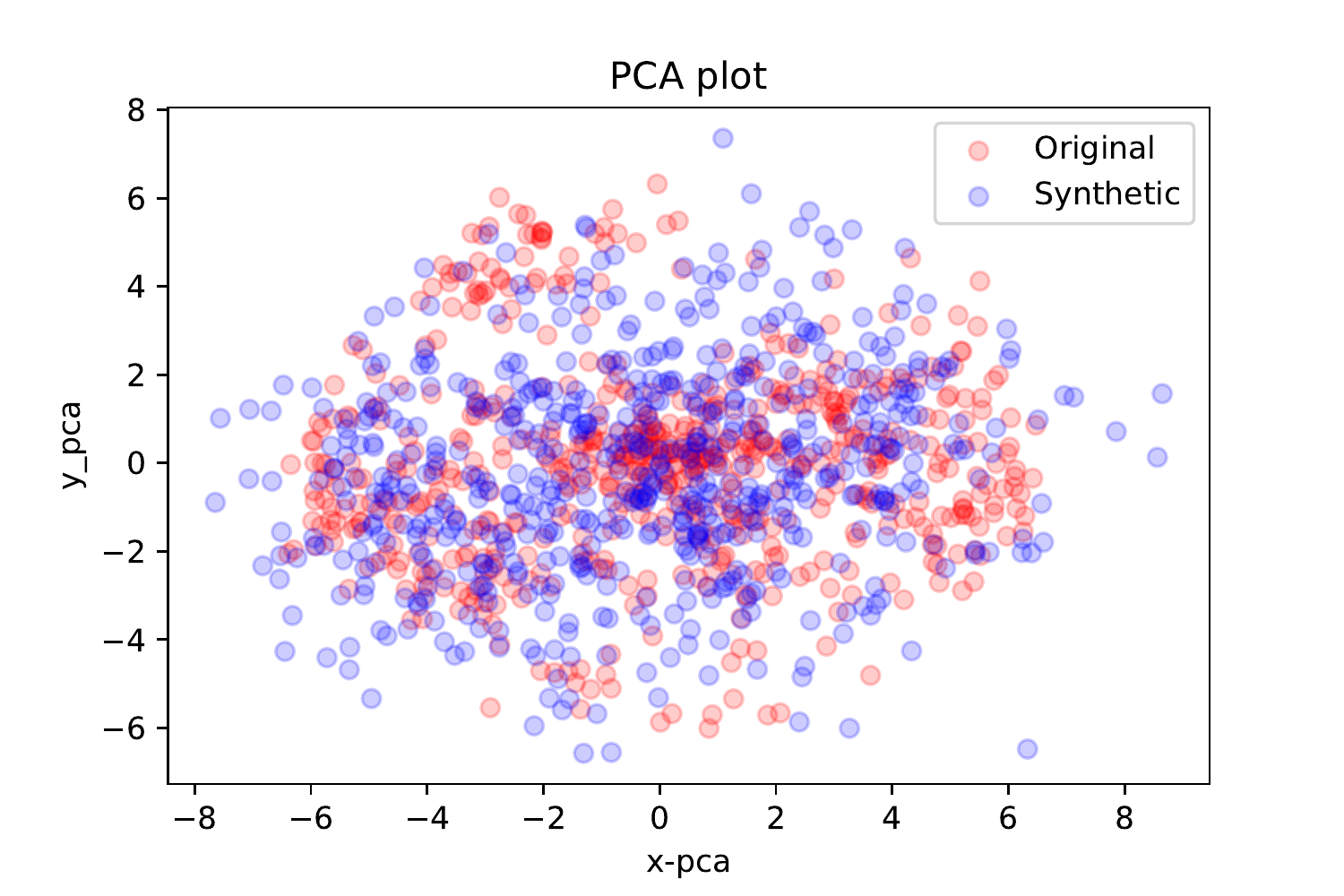}
    \end{subfigure}
    \begin{subfigure}{0.245\textwidth}
        \centering
        \includegraphics[width=\columnwidth]{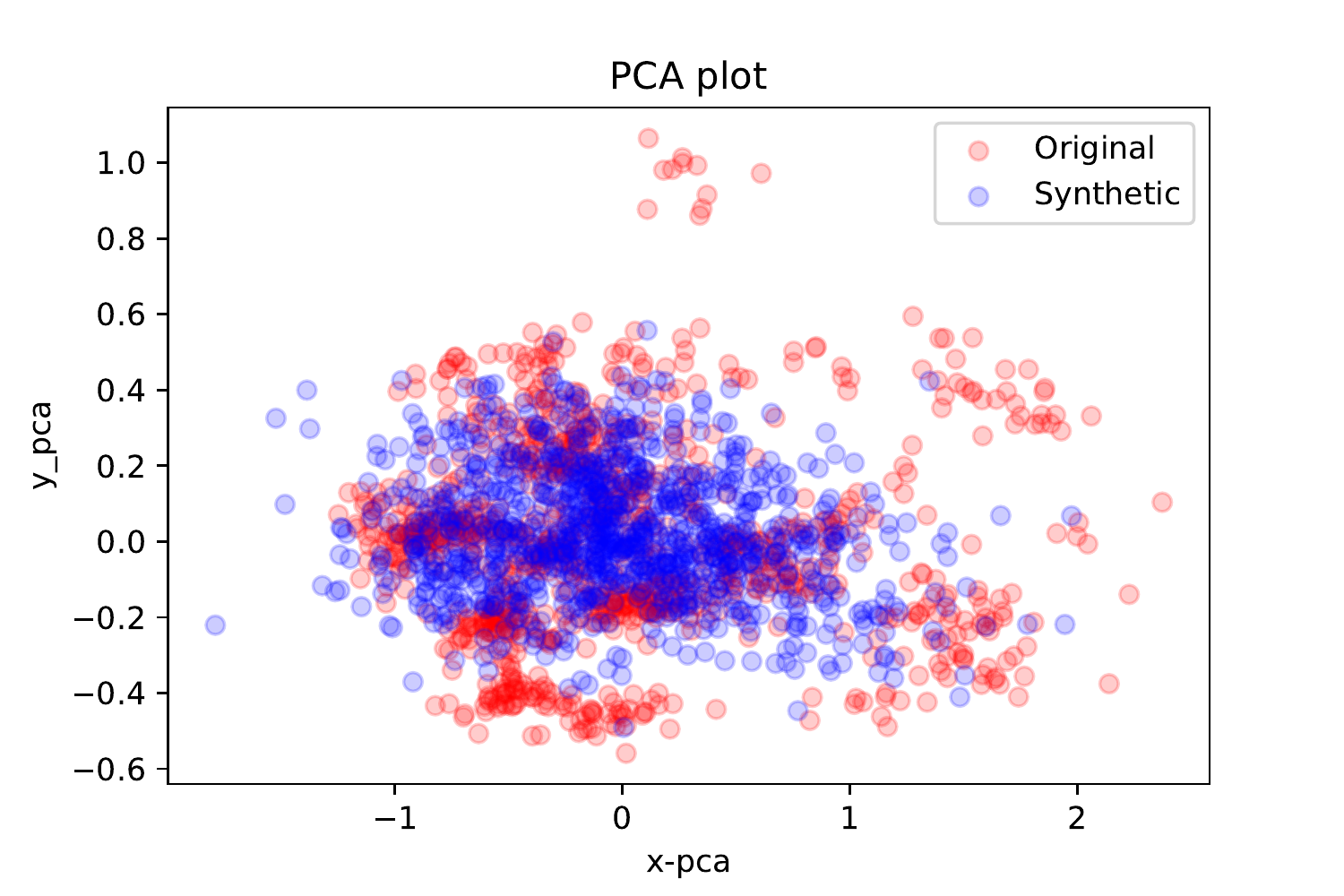}
    \end{subfigure}
    \begin{subfigure}{0.245\textwidth}  
        \centering 
        \includegraphics[width=\columnwidth]{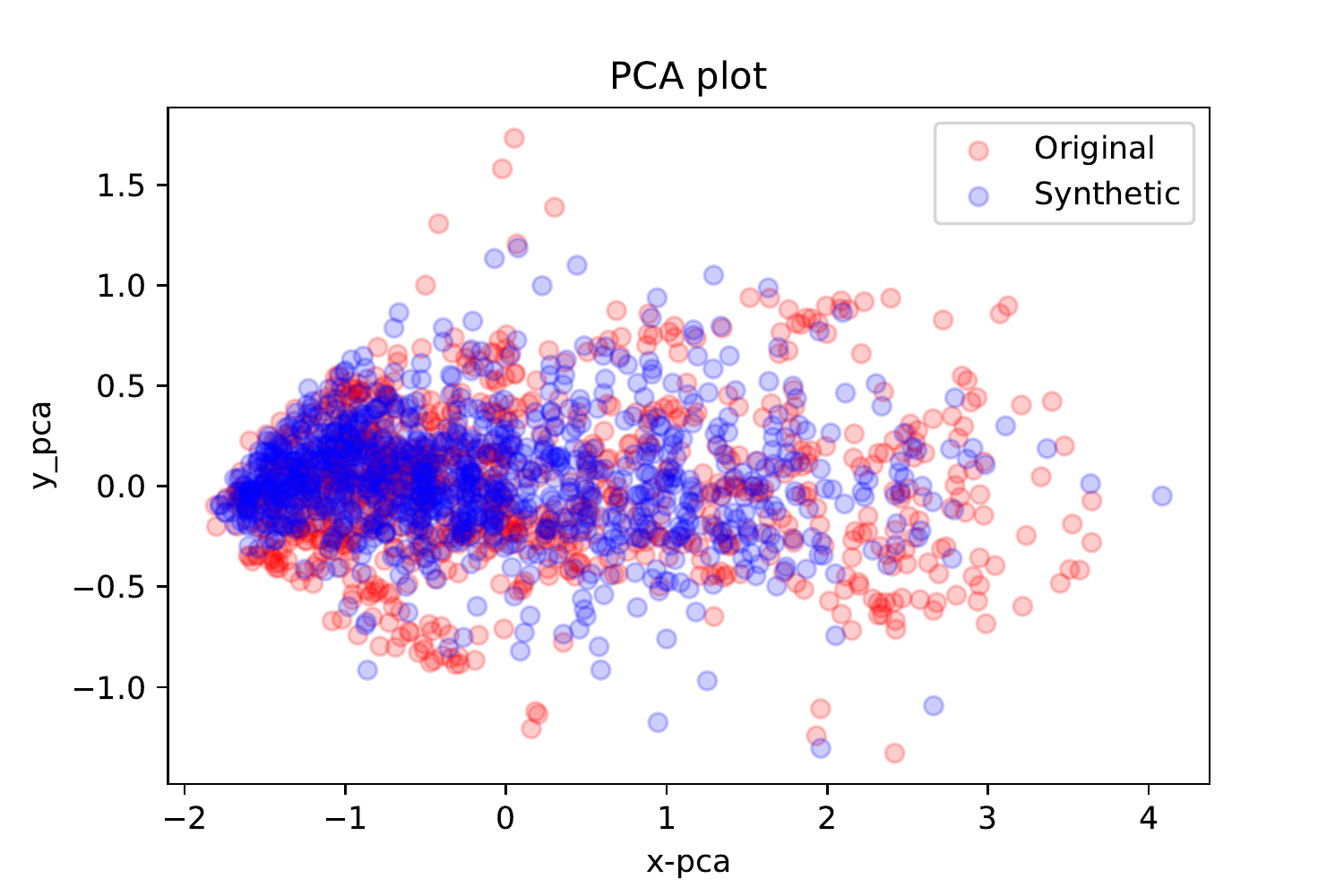}
    \end{subfigure}
    
    \begin{subfigure}{0.245\textwidth}   
        \centering 
        \includegraphics[width=\columnwidth]{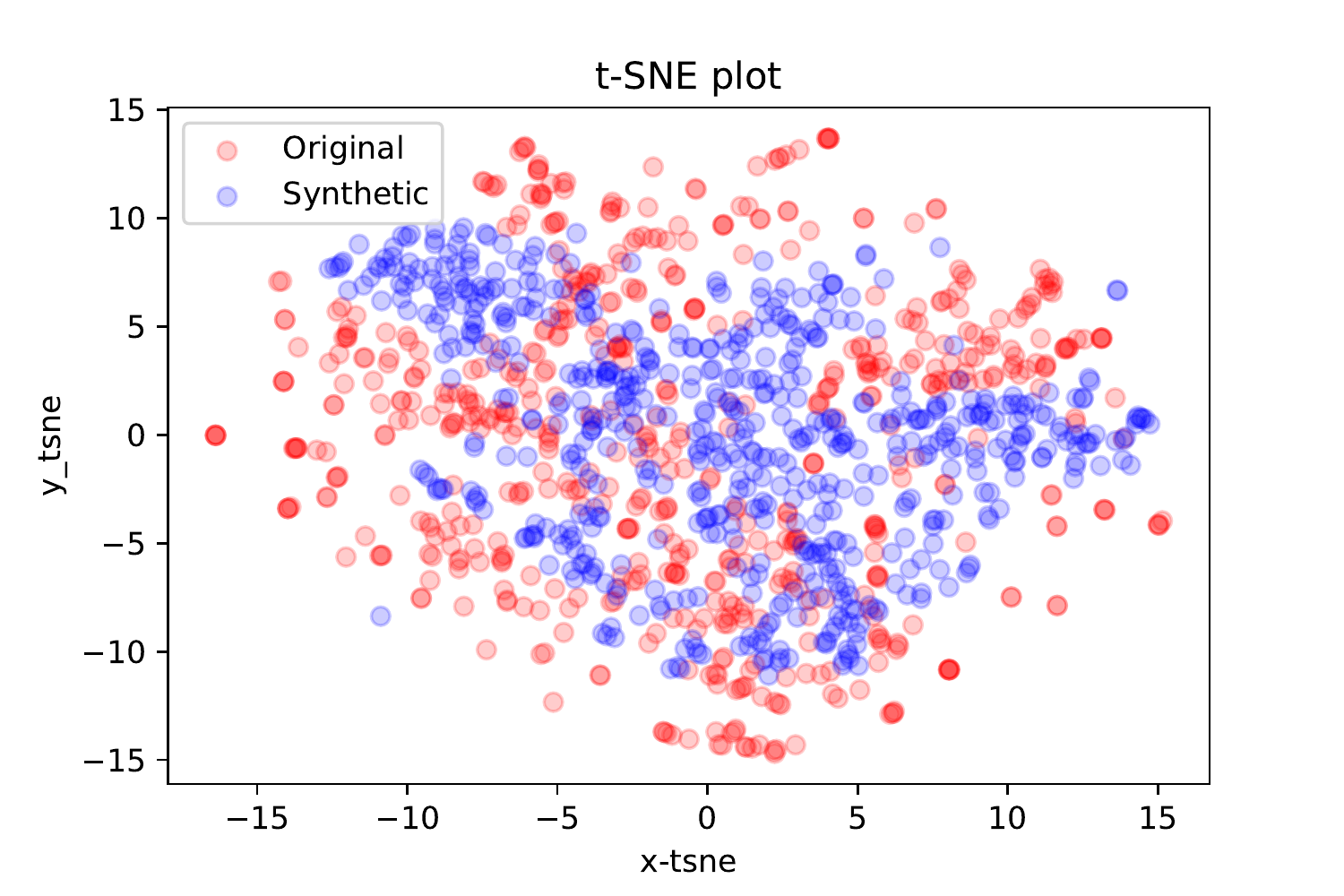}
        \caption{Jumping}
    \end{subfigure}
    \begin{subfigure}{0.245\textwidth}   
        \centering 
        \includegraphics[width=\columnwidth]{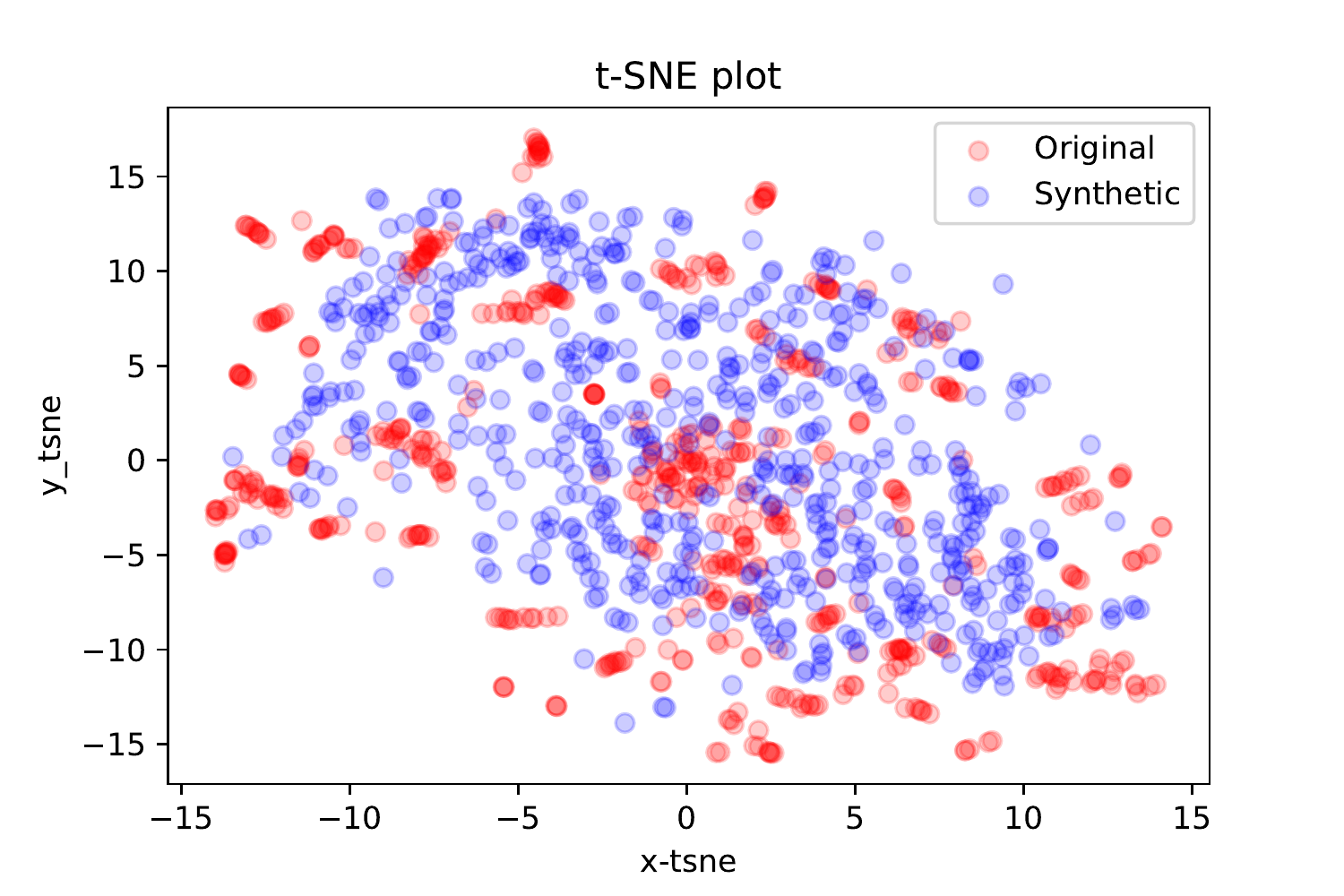}
        \caption{Running}
    \end{subfigure}
    \begin{subfigure}{0.245\textwidth}   
        \centering 
        \includegraphics[width=\columnwidth]{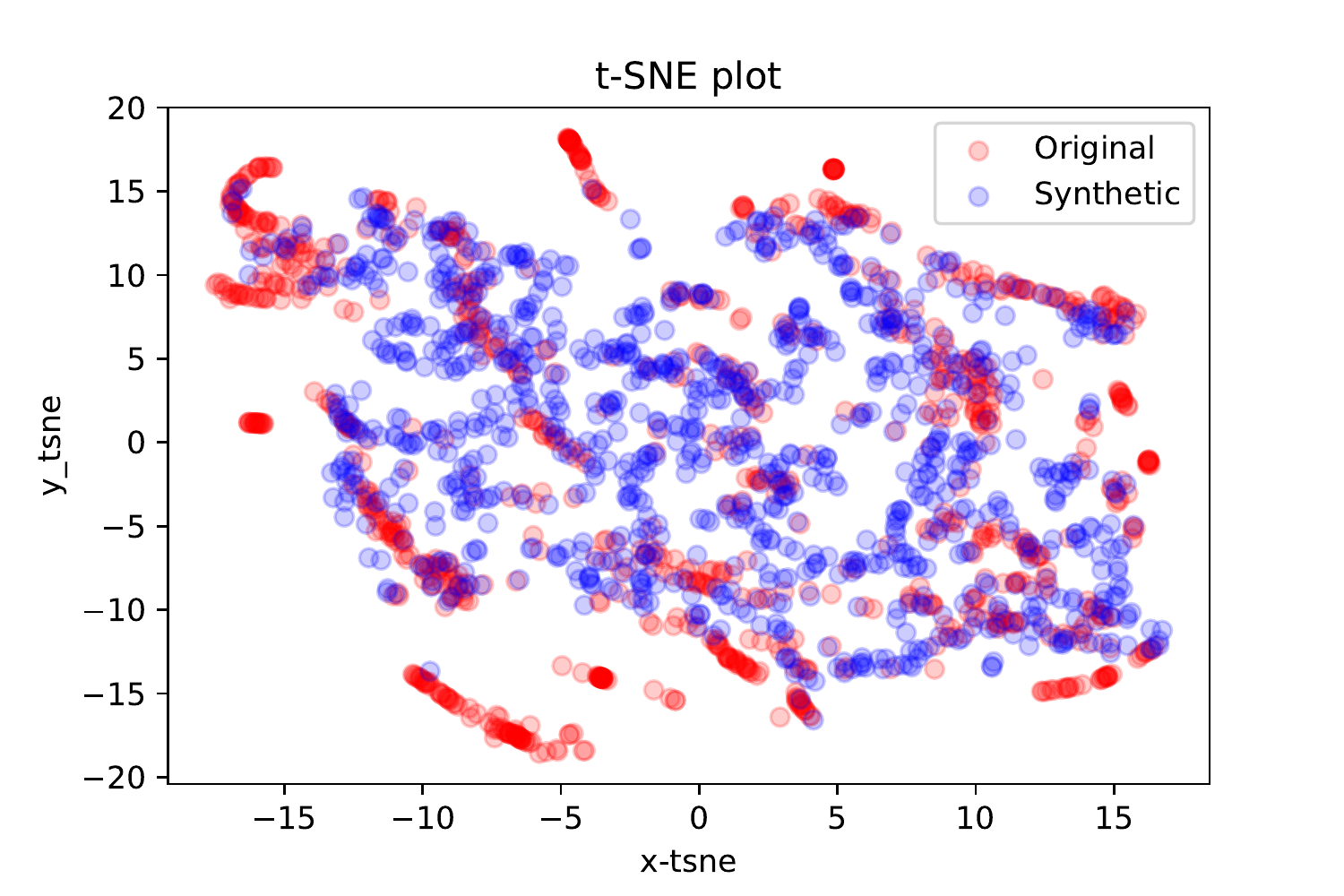}
        \caption{Normal ECG}
    \end{subfigure}
    \begin{subfigure}{0.245\textwidth}   
        \centering 
        \includegraphics[width=\columnwidth]{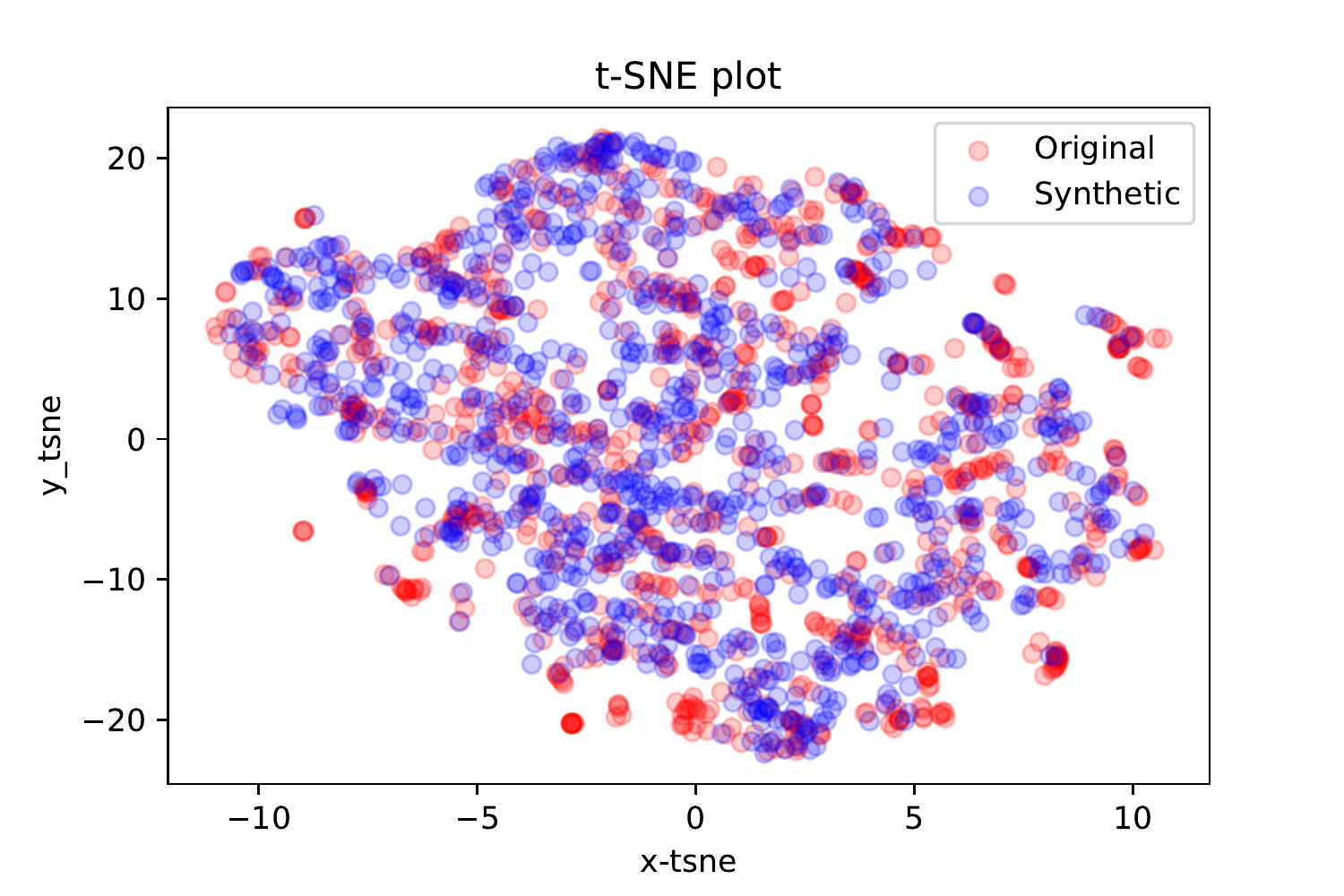}
        \caption{Abnormal ECG}
    \end{subfigure}
    
    \caption{The PCA and t-SNE test for real and synthetic data generated by TTS-GAN.}
    \label{fig:smallvisual}
    
\end{figure*}

\noindent \textbf{Visualizations with PCA and t-SNE:}
To further illustrate the similarity between the real data and synthetic data, we plot visualization example graphs of data point distributions mapped to two dimensions using PCA and t-SNE in Figure~\ref{fig:smallvisual}. In these plots, red dots denote original data, and blue dots denote synthetic data generated by TTS-GAN. Again, we notice a similar distribution pattern between real and synthetic data. 


\noindent \textbf{Similarity scores:}
To quantitatively compare the similarity of the real and generated sequences, we defined two similarity scores, average cosine similarity ($avg\_cos\_sim$) and average Jensen-Shannon distance ($avg\_jen\_dis$). The detailed definition of these similarity metrics is given in Appendix~\ref{sec:similarityScores}. We first extract 7 well-known signal features from each signal channel $C$, to form a $7 \times C$ dimensional feature vector for each sequence. The $avg\_cos\_sim$ measures the average cosine similarity among all real signals and synthetic signals of the same class. Values closer to 1 indicate high similarity between two feature vectors. The Jensen–Shannon distance is a method of measuring the similarity between two probability-like distributions. We consider each extracted feature to be a normally distributed array of values and compute the Jensen-Shannon distance for corresponding features between real and synthetic feature vectors. The $avg\_jen\_dis$ is the average of all feature vector distances. A value closer to zero means a pair of signals have a small distance from each other and thus share similar distributions. As it can be observed from the experimental results shown in Table~\ref{tbl:similarityscores}, synthetic samples show a high average cosine similarity and low Jensen–Shannon distance for different signal classes. In addition, TTS-GAN wins against Time-GAN in 7 out of 10 cases.

\begin{table}[]
\centering
\begin{tabular}{c|c|ccccc}
\hline
Model Name & Similarity Score & Sinusoidal      & Jumping         & Running         & Normal          & Abnormal        \\ \hline
TTS-GAN    & avg\_cos\_sim    & \textbf{0.9936} & \textbf{0.9982} & 0.9988          & 0.9855          & \textbf{0.9768} \\
           & avg\_jen\_dis    & \textbf{0.0980} & \textbf{0.0870} & 0.0497          & \textbf{0.1861} & \textbf{0.2911} \\
Time-GAN   & avg\_cos\_sim    & 0.9935          & 0.9980          & \textbf{0.9989} & \textbf{0.9878} & 0.9719          \\
           & avg\_jen\_dis    & 0.1226          & 0.0924          & \textbf{0.0470} & 0.1883          & 0.3354          \\ \hline
\end{tabular}
    \caption{The similarity scores between real data and synthetic data of 5 different datasets. $avg\_cos\_sim$, the bigger the better. $avg\_jen\_dis$, the smaller the better. Bold texts identify better results.}
    \label{tbl:similarityscores}
    
\vspace{-5mm}
\end{table}


\section{Conclusions}
\label{sec:conclude}

In this work, we build a transformer-based GAN model (TTS-GAN) that is able to generate multi-dimensional time-series data of various lengths. A visual comparison of the raw signal patterns as well as data point distributions mapped in two dimensions show the similarity of the original data and the synthetic data. Two similarity scores are also used to quantitatively further verify the fidelity of the synthetic data. Overall, the experimental results demonstrate the viability of TTS-GAN as a generator of realistic time-series, when trained on real samples.

%
%
%
\bibliographystyle{ieeetr}
\bibliography{camera_ready}

\begin{thebibliography}{10}

\bibitem{goodfellow2014generative}
I.~Goodfellow, J.~Pouget-Abadie, M.~Mirza, B.~Xu, D.~Warde-Farley, S.~Ozair,
  A.~Courville, and Y.~Bengio, ``Generative adversarial nets,'' {\em Advances
  in neural information processing systems}, vol.~27, 2014.

\bibitem{ledig2017photo}
C.~Ledig, L.~Theis, F.~Husz{\'a}r, J.~Caballero, A.~Cunningham, A.~Acosta,
  A.~Aitken, A.~Tejani, J.~Totz, Z.~Wang, {\em et~al.}, ``Photo-realistic
  single image super-resolution using a generative adversarial network,'' in
  {\em Proceedings of the IEEE conference on computer vision and pattern
  recognition}, pp.~4681--4690, 2017.

\bibitem{bousmalis2017unsupervised}
K.~Bousmalis, N.~Silberman, D.~Dohan, D.~Erhan, and D.~Krishnan, ``Unsupervised
  pixel-level domain adaptation with generative adversarial networks,'' in {\em
  Proceedings of the IEEE conference on computer vision and pattern
  recognition}, pp.~3722--3731, 2017.

\bibitem{zhang2017stackgan}
H.~Zhang, T.~Xu, H.~Li, S.~Zhang, X.~Wang, X.~Huang, and D.~N. Metaxas,
  ``Stackgan: Text to photo-realistic image synthesis with stacked generative
  adversarial networks,'' in {\em Proceedings of the IEEE international
  conference on computer vision}, pp.~5907--5915, 2017.

\bibitem{brophy2021generative}
E.~Brophy, Z.~Wang, Q.~She, and T.~Ward, ``Generative adversarial networks in
  time series: A survey and taxonomy,'' {\em arXiv preprint arXiv:2107.11098},
  2021.

\bibitem{vaswani2017attention}
A.~Vaswani, N.~Shazeer, N.~Parmar, J.~Uszkoreit, L.~Jones, A.~N. Gomez,
  {\L}.~Kaiser, and I.~Polosukhin, ``Attention is all you need,'' in {\em
  Advances in neural information processing systems}, pp.~5998--6008, 2017.

\bibitem{dosovitskiy2020image}
A.~Dosovitskiy, L.~Beyer, A.~Kolesnikov, D.~Weissenborn, X.~Zhai,
  T.~Unterthiner, M.~Dehghani, M.~Minderer, G.~Heigold, S.~Gelly, {\em et~al.},
  ``An image is worth 16x16 words: Transformers for image recognition at
  scale,'' {\em arXiv preprint arXiv:2010.11929}, 2020.

\bibitem{devlin2018bert}
J.~Devlin, M.-W. Chang, K.~Lee, and K.~Toutanova, ``Bert: Pre-training of deep
  bidirectional transformers for language understanding,'' {\em arXiv preprint
  arXiv:1810.04805}, 2018.

\bibitem{lu2021pretrained}
K.~Lu, A.~Grover, P.~Abbeel, and I.~Mordatch, ``Pretrained transformers as
  universal computation engines,'' {\em arXiv preprint arXiv:2103.05247}, 2021.

\bibitem{jiang2021transgan}
Y.~Jiang, S.~Chang, and Z.~Wang, ``Transgan: Two pure transformers can make one
  strong gan, and that can scale up,'' in {\em Thirty-Fifth Conference on
  Neural Information Processing Systems}, 2021.

\bibitem{diao2021tilgan}
S.~Diao, X.~Shen, K.~Shum, Y.~Song, and T.~Zhang, ``Tilgan: Transformer-based
  implicit latent gan for diverse and coherent text generation,'' in {\em
  Findings of the Association for Computational Linguistics: ACL-IJCNLP 2021},
  pp.~4844--4858, 2021.

\bibitem{esteban2017real}
C.~Esteban, S.~L. Hyland, and G.~R{\"a}tsch, ``Real-valued (medical) time
  series generation with recurrent conditional gans,'' {\em arXiv preprint
  arXiv:1706.02633}, 2017.

\bibitem{yoon2019time}
J.~Yoon, D.~Jarrett, and M.~Van~der Schaar, ``Time-series generative
  adversarial networks,'' 2019.

\bibitem{ni2020conditional}
H.~Ni, L.~Szpruch, M.~Wiese, S.~Liao, and B.~Xiao, ``Conditional
  sig-wasserstein gans for time series generation,'' {\em arXiv preprint
  arXiv:2006.05421}, 2020.

\bibitem{wold1987principal}
S.~Wold, K.~Esbensen, and P.~Geladi, ``Principal component analysis,'' {\em
  Chemometrics and intelligent laboratory systems}, vol.~2, no.~1-3,
  pp.~37--52, 1987.

\bibitem{van2008visualizing}
L.~Van~der Maaten and G.~Hinton, ``Visualizing data using t-sne.,'' {\em
  Journal of machine learning research}, vol.~9, no.~11, 2008.

\bibitem{ratliff2013characterization}
L.~J. Ratliff, S.~A. Burden, and S.~S. Sastry, ``Characterization and
  computation of local nash equilibria in continuous games,'' in {\em 2013 51st
  Annual Allerton Conference on Communication, Control, and Computing
  (Allerton)}, pp.~917--924, IEEE, 2013.

\bibitem{goodfellow2020generative}
I.~Goodfellow, J.~Pouget-Abadie, M.~Mirza, B.~Xu, D.~Warde-Farley, S.~Ozair,
  A.~Courville, and Y.~Bengio, ``Generative adversarial networks,'' {\em
  Communications of the ACM}, vol.~63, no.~11, pp.~139--144, 2020.

\bibitem{huang2017beyond}
R.~Huang, S.~Zhang, T.~Li, and R.~He, ``Beyond face rotation: Global and local
  perception gan for photorealistic and identity preserving frontal view
  synthesis,'' in {\em Proceedings of the IEEE international conference on
  computer vision}, pp.~2439--2448, 2017.

\bibitem{app7101101}
D.~Micucci, M.~Mobilio, and P.~Napoletano, ``Unimib shar: A dataset for human
  activity recognition using acceleration data from smartphones,'' {\em Applied
  Sciences}, vol.~7, no.~10, 2017.

\bibitem{bousseljot1995nutzung}
R.~Bousseljot, D.~Kreiseler, and A.~Schnabel, ``Nutzung der ekg-signaldatenbank
  cardiodat der ptb {\"u}ber das internet,'' 1995.

\bibitem{goldberger2000physiobank}
A.~L. Goldberger, L.~A. Amaral, L.~Glass, J.~M. Hausdorff, P.~C. Ivanov, R.~G.
  Mark, J.~E. Mietus, G.~B. Moody, C.-K. Peng, and H.~E. Stanley, ``Physiobank,
  physiotoolkit, and physionet: components of a new research resource for
  complex physiologic signals,'' {\em circulation}, vol.~101, no.~23,
  pp.~e215--e220, 2000.

\bibitem{mao2017least}
X.~Mao, Q.~Li, H.~Xie, R.~Y. Lau, Z.~Wang, and S.~Paul~Smolley, ``Least squares
  generative adversarial networks,'' in {\em Proceedings of the IEEE
  international conference on computer vision}, pp.~2794--2802, 2017.

\end{thebibliography}

\appendix             

\section{Appendix 1: Training Details}
We conduct all experiments on an Intel server with a 3.40GHz CPU, 377GB RAM memory and 2 Nvidia 1080 GPUs. 
For all datasets, the synthetic data are generated by a generator that takes random vectors of size $(100, 1)$ as inputs. The transformer blocks in the generator and discriminator are both repeated three times. We adopt a learning rate of $1e-4$ for the generator and $3e-4$ for the discriminator. We follow the setting of LSGAN~\cite{mao2017least} and use loss function described in section~\ref{sec:loss} to update model parameters. An Adam optimizer with $\beta_1 = 0.9$ and $\beta_2 = 0.999$, and a batch size of 32 for both generator and discriminator, are used for all experiments.

\section{Appendix 2: Similarity Scores}
\label{sec:similarityScores}
\noindent \textbf{Feature extraction}
We extract several meaningful features from each input data sequence. They are the median, mean, standard deviation, variance, root mean square, maximum, and minimum values of each input sequence. 
Suppose we compute $m$ features from all channels of each sequence and get a feature vector with the format $f = <feature_1, feature_2, ..., feature_m>$. 

\noindent \textbf{Average Cosine Similarity}
For each pair of real signal feature vector $f_a$ and synthetic signal feature vector $f_b$, the vector has the size $m$, we can compute its cosine similarity as:
\begin{equation*}
    cos\_sim_{ab} = \frac{f_a\cdot f_b}{\left \| f_a \right \|\left \| f_b \right \|} = 
    \frac{\sum_{i=1}^{m}f_{ai} f_{bi}}{\sqrt{\sum_{i=1}^{m}f_{ai}^{2}}{\sqrt{\sum_{i=1}^{m}f_{bi}^{2}}}}
\vspace{-2mm}
\end{equation*}

The average cosine similarity score is the average of each cosine similarity between pairs of feature vectors corresponding to real and synthetic signals of the same class. The average cosine similarity is computed as follows, where $n$ the total number of signals:
\begin{equation*}
    avg\_cos\_sim = \frac{1}{n} \sum_{i=1}^{n}cos\_sim_i
\vspace{-2mm}
\end{equation*}

\noindent \textbf{Average Jensen-Shannon distance}
The average jensen-shannon distance is the average of jensen-shannon distance between each feature from real signals and synthetic signals. 
For each pair of real signal feature $f_{i\_real}$ and synthetic signal feature $f_{i\_syn}$, we can compute its jensen-shannon distance as:
\begin{equation*}
jen\_sim_i = \sqrt{\frac{D(f_{i\_real} || m)+D(f_{i\_syn} || m)}{2}}
\vspace{-2mm}
\end{equation*}

Where $m$ is the pointwise mean of $f_{i\_real}$ and $f_{i\_syn}$ and $D$ is the Kullback-Leibler divergence. The average jensens-shannon distance is computed as:
\begin{equation*}
    avg\_jen\_dis = \sum_{i=1}^{m}jen\_sim_i
\vspace{-2mm}
\end{equation*}

\end{document}